\pdfoutput=1

\documentclass[11pt]{article}

\usepackage[]{acl}

\usepackage{times}
\usepackage{latexsym}
\usepackage{booktabs,colortbl}

\usepackage[T1]{fontenc}

\usepackage[utf8]{inputenc}

\usepackage{microtype}

\usepackage{inconsolata}

\usepackage{multirow}
\usepackage{amsmath}
\usepackage{booktabs}
\usepackage{graphicx}
\usepackage{cleveref}
\usepackage{xspace}

\usepackage{caption}
\usepackage{subcaption}
\usepackage{colortbl}
\usepackage{color}
\usepackage{xcolor}
\crefname{figure}{Figure}{Figures}
\crefname{table}{Table}{Tables}
\crefname{appendix}{Appendix}{Appendices}

\definecolor{gr}{RGB}{169,169,196} 
\newcommand{\gr}[0]{\color{gr}}

\title{Transferring Fairness using Multi-Task Learning \\ with Limited Demographic Information}

\author{Carlos Aguirre \and Mark Dredze \\
  Johns Hopkins University \\
  \texttt{caguirre@cs.jhu.edu} \and \texttt{mdredze@cs.jhu.edu}\\}

\begin{document}
\maketitle
\begin{abstract}
Training supervised machine learning systems with a fairness loss can improve prediction fairness across different demographic groups. 
However, doing so requires demographic annotations for training data, without which we cannot produce debiased classifiers for most tasks.
Drawing inspiration from transfer learning methods, we investigate whether we can utilize demographic data from a related task to improve the fairness of a target task.
We adapt a single-task fairness loss to a multi-task setting to exploit demographic labels from a related task in debiasing a target task, and demonstrate that demographic fairness objectives transfer fairness within a multi-task framework. Additionally, we show that this approach enables intersectional fairness by transferring between two datasets with different single-axis demographics.
We explore different data domains to show how our loss can improve fairness domains and tasks.
\end{abstract}

\section{Introduction}

\begin{figure}[ht]
    \includegraphics[width=\columnwidth,keepaspectratio]{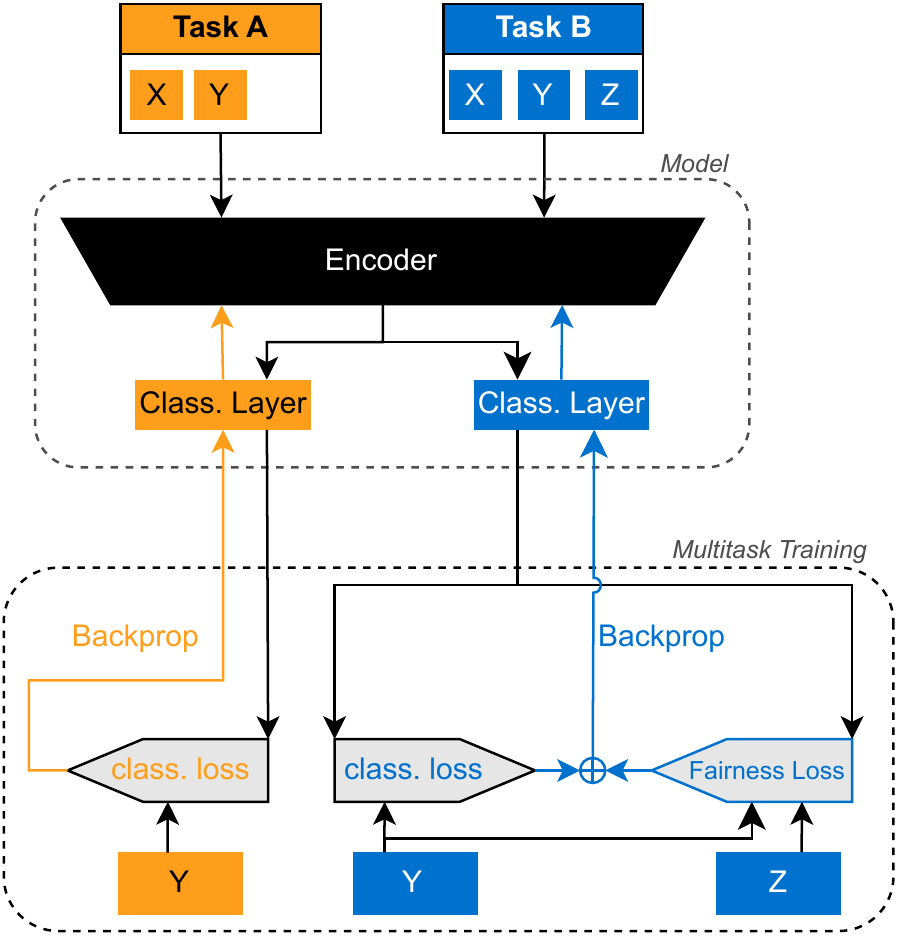}
    \caption{Our approach, \textit{MTL fair}, a multitask method to utilize an auxiliary task (B) to train a fair model for a task (A) without demographic annotations.}
    
    \label{fig:MTL-fair}
\end{figure}

Machine learning models can have disparate performance on specific subpopulations even when they have relatively high performance overall, which can mask poor performance for smaller subpopulations.
To alleviate disparate performance and biased model behavior, a variety of techniques can make for fairer AI systems, such as additional training objectives to debias models.
These training objectives require example metadata, such as author demographics, to influence the loss toward fairer model behavior.
Unfortunately, training set demographic metadata is often unavailable, thus creating a barrier to training fair systems.

Transfer learning is a general strategy for learning with limited or no training labels, where annotations from one task are used to train a model in a related task. Multi-task learning (MTL) utilizes transfer learning to jointly train a model over several related tasks. We draw inspiration from MTL methods and ask, \textit{can MTL transfer demographic fairness between related tasks? }
Suppose we have target labels for two tasks A and B, but demographic labels only for task A; can we transfer fairness learned from task A to task B?
We adapt existing MTL and fairness loss methods to achieve the goal of demographic fairness transfer.
\Cref{fig:MTL-fair} shows a representation of our method to achieve model fairness given demographic annotations for only one task.

The success of this approach can be adapted to address a limitation in current demographic fairness methods: intersectional fairness. 
Intersectional fairness means that fairness conditions hold across cross-products of orthogonal attributes and not just within a single attribute. 
\citet{crenshaw1989demarginalizing} introduced the term \textit{intersectionality} in the legal field\footnote{The idea can be found in prior sources \citep{truth1851ain}, as described in \citet{costanza2020design}.} to describe how anti-discrimination laws failed to protect Black women workers, as employers avoided charges of discrimination by hiring enough Black men and White women to satisfy the single-identity clauses.
Similarly, early work in machine learning found biases in vision models at the intersection of gender and skin color \citep{buolamwini2018gender}, where facial recognition models performed worse for Black women.
Current methods cannot enforce intersectional fairness unless we have annotations for both attributes on the same instances. This high bar for training data further exacerbates data scarcity since most datasets with demographic attributes only consider single-axis attributes (e.g. race or gender alone.)
Therefore, we use our MTL approach to 
produce an intersectionally fair model for two tasks (gender \textit{and} race) from a dataset from different single-axis demographic annotations for each task (i.e. gender \textit{or} race).

Finally, we explore how the relationship between tasks enables fairness transfer by conducting experiments with different tasks in two domains (clinical and social media) and evaluate the fairness transfer between tasks within and across domains.

We summarize our contributions as follows:
\begin{itemize}
    \setlength\itemsep{.01em}
    \setlength{\parskip}{0pt}
    \setlength{\parsep}{0pt}
    \item We transfer fairness across tasks by adapting single-task fairness losses to multi-task settings.
    \item We enable intersectional fairness by leveraging two tasks with single-axis demographic attributes using a multi-task fairness loss.
    \item We explore the relationship between task similarity and fairness generalization.
\end{itemize}

\section{Methods}

We begin by describing the learning setting shown in \cref{fig:MTL-fair}.
Let us assume we desire an unbiased model for task A for which we have input text (X) and associated labels (Y), but no demographic attributes.
Instead, we have demographic data for task B, a task related to but distinct from A. 
Since there exist similarities between tasks A and B, we wish to utilize the demographic attributes (Z) available for task B to obtain a fair classifier for task A. Specifically, by using multi-task training to jointly train a model with both tasks A and B, with an added fairness loss supported by task B alone, we hope to produce a fair model for task A.

Employing a similar idea, we generalize our approach to intersectional fairness.
We want to train classifiers for both tasks A and B, which consist of text data and target labels. We have demographic attributes for both A and B, but they are {\em different} attributes for each task, e.g. task A has gender attributes and task B has race attributes. 
Since neither task has both attributes, we are unable to utilize an intersectional fairness loss to the tasks individually.
Therefore, we propose a multi-task objective to combine attributes from both tasks to obtain intersectional fairness.

This section introduces our fairness definitions and losses, provides formal definitions of our training objectives and describes our training procedure.

\subsection{Fairness Loss and Definition}
\label{sec:methods-fair-loss}

\begin{table*}
\centering
\resizebox{2\columnwidth}{!}{
    \begin{tabular}{cl@{ }l}
    \toprule
    \multicolumn{1}{c}{Fairness loss} & \multicolumn{2}{c}{Objective} \\
    \cmidrule(lr){1-1} \cmidrule(lr){2-3} 
        single task   & \( \min_{\theta} f(X; \theta) \) & \(\overset{\Delta}{=} \frac{1}{N} \sum_{i=1}^{N} \mathcal{L}(x_i; \theta) + \lambda[\max(0, \epsilon(X;\theta) - \epsilon_t)] \)    \\\\
        \multirow{2}{*}{MTL}   & \( \min_{\theta} f(A;B; \theta) \) & \( \overset{\Delta}{=} 
    \frac{1}{|A||B|} \sum_{i=1}^{|A|} \sum_{j=1}^{|B|} \mathcal{L}(x_{a,i}; [\theta_s \cup \theta_a]) 
     \) \\ & & \(
    + \mathcal{L}(x_{b,i}; [\theta_s \cup \theta_b])  
    + \lambda[\max(0, \epsilon(B;[\theta_s \cup \theta_b]) - \epsilon_t)]\)      \\ \\
        \multirow{2}{*}{\begin{tabular}[c]{@{}c@{}}MTL \\ intersectional\end{tabular}} & \( \min_{\theta} f(A;B; \theta) \) & \( \overset{\Delta}{=} \frac{1}{|A||B|} \sum_{i=1}^{|A|} \sum_{j=1}^{|B|} \mathcal{L}(x_{a,i}; [\theta_s \cup \theta_a])  +  \lambda[\max(0, \epsilon(A;[\theta_s \cup \theta_a]) - \epsilon_t)] 
     \) \\ & & \(+ \mathcal{L}(x_{b,i}; [\theta_s \cup \theta_b]) 
    + \lambda[\max(0, \epsilon(B;[\theta_s \cup \theta_b]) - \epsilon_t)] \) \\
    \bottomrule
    \end{tabular}}
    \caption{\label{tab:fair-losses}
    Objectives for adding fairness losses in single task, MTL and MTL intersectional cases.
    }
\end{table*}

We select a fairness definition that supports intersectionality and that is differentiable so that it can be included in model training. 
We use $\epsilon$-Differential Equalized Odds ($\epsilon$-DEO), a variant of $\epsilon$-DF \citep{foulds2020intersectional}, that applies the equalized odds objective, with the goal of equalizing recall and specificity rates across demographic groups \citep{barocas-hardt-narayanan} and intersectional subgroups, and that is learnable and differentiable.
We apply \textit{equalized odds} on the $\epsilon$-DF framework and call it  $\epsilon$-Differential Equalized Odds ($\epsilon$-DEO).
Formally, let $s_1, ... , s_p$ be discrete-valued demographic attributes, and  $z = s_1 \times s_2 \times ... \times s_p$ the intersectional groups. A model $M(X)$ satisfies $\epsilon$-DEO with respect to $z$ if for all $x$, $\hat{y} \in \textrm{Range}(M)$ and $y \in \textrm{Range}(M)$,
\begin{equation}
    e^{-\epsilon} \leq \frac{Pr(M(x) = \hat{y} | \zeta_i, y)}{Pr(M(x) = \hat{y} | \zeta_j, y)} \leq e^{\epsilon},
\end{equation}
for all $(\zeta_i, \zeta_j) \in z \times z$ where $Pr(\zeta_i) > 0$, $Pr(\zeta_j) > 0$; smaller $\epsilon$ is better, with $\epsilon$ = 0 for perfect fairness.
Under $\epsilon$-DEO, perfect fairness results from a classifier with the same recall and specificity rates across intersectional groups of demographic attributes.
Utilizing the equalized odds objective is important--as opposed to others, e.g. \textit{demographic parity}--because it avoids limitations that arise when the labels are correlated with demographic variables, which is the case in many real-world problems and the datasets used in our experiments, e.g. the clinical datasets used in our paper \citep{hardt2016equality}. More information about the fairness defintions is provided in \Cref{apx:fairness-definition}.

The standard approach to incorporating fairness metrics into learning objectives uses an additive term. For example, for a deep neural network classifier $M(X)$ with parameters $\theta$, we obtain the \textit{single task} equation in ~\cref{tab:fair-losses}, where $\epsilon(X;\theta)$ is the $\epsilon$-DEO measure for the classifier, $\epsilon_t$ is the desired base fairness (in our experiments 0), and $\lambda$ is a hyper-parameter that trades between prediction loss and fairness \citep{foulds2020intersectional}.
Since the fairness term is differentiable, the model can be trained using stochastic gradient descent on the objective via backpropagation and automatic differentiation.
A \textit{burn-in} period and stochastic approximation-based update are adopted following \citet{foulds2020intersectional}.
One optimization challenge that emerges from incorporating fairness is instability due to the representativeness of the mini-batches: a diverse set of examples is needed on which the fairness loss can be meaningfully measured.
Following prior work \citep{foulds2020intersectional}, we use a stochastic approximation-based update for $\epsilon(X;\theta)$ by estimating mini-batch noisy expected counts per intersecting demographic group with a hyperparameter $\rho$, $\tilde{\mathcal{N}}_{t} = (1-\rho)\tilde{\mathcal{N}}_{t-1} + \rho \mathcal{N}_{t}$, where $\tilde{\mathcal{N}_{t}}$ is the approximated count at time $t$ and $\mathcal{N}_{t}$ is the actual count. Thus $\rho$ controls the smoothness of the approximation of the demographic counts in mini-batches.

\subsection{MTL fairness}
\label{sec:methods-mtl-fairness}

We train a model jointly on tasks A and B with a fairness loss applied only to task B, as seen in \cref{fig:MTL-fair} (\textit{MTL fair}.)
The MTL training will optimize the shared model parameters (the encoder) to exploit task similarities and improve fairness in task A based on the fairness constraints of task B.

Assume we have a target task $A$ with training instances of input features $x_a$ and task labels $y_a$, and an auxiliary task $B$, with training instances of input features $x_b$, task labels $y_b$ and demographic attributes $z_b$.
Adding the fairness loss with respect to task $B$ in a multi-task objective of a DNN-based classifier $M(X)$ with shared parameters $\theta_s$, task $A$-specific parameters $\theta_a$ and task $B$-specific parameters $\theta_b$, where $\theta = (\theta_s \cup \theta_a \cup \theta_b)$ becomes \textit{MTL} equation in \cref{tab:fair-losses},
where $\epsilon(B;[\theta_s \cup \theta_b])$ is the $\epsilon$-DEO measure for the classifier on task $B$.
Notably, $\epsilon(B;[\theta_s \cup \theta_b])$ is applied to both task-specific and shared parameters.

\subsection{Intersectionality}
\label{sec:methods-mtl-intersectional-fairness}
We formalize the problem of intersectional fairness across tasks using the $\epsilon$-DEO loss across both tasks using
MTL training with two fairness losses, one for each task.

Assume we have a target task $A$, with training instances of input features $x_a$, task labels $y_a$, and demographic attributes $w_a$, and an auxiliary task $B$ with training instances of input features $x_b$, task labels $y_b$ and demographic attributes $w_b$.
We seek an intersectionally fair classifier on both tasks with respect to $z=w_a \times w_b$.
Adding the fairness loss in a multi-task objective of a DNN-based classifier $M(X)$ with shared parameters $\theta_s$, task $A$-specific parameters $\theta_a$ and task $B$-specific parameters $\theta_b$, where $\theta = (\theta_s \cup \theta_a \cup \theta_b)$ \textit{MTL intersectional} equation in \cref{tab:fair-losses},
where $\epsilon(A;[\theta_s \cup \theta_a])$ and $\epsilon(B;[\theta_s \cup \theta_b])$ are the $\epsilon$-DEO measure for the classifier on task $A$ and $B$ respectively.
Notably, both losses update the shared parameters $\theta_s$.

\section{Data}
\label{sec:data}

While our method can transfer demographic fairness from one task to another when only one has demographic information, we 
need a dataset with multiple demographic attributes and attributes for each task to test intersectional fairness.
We select datasets in varied domains: clinical text records, online reviews, and social media (\cref{tab:data-summary}.)
\Cref{apx:data-details} gives a detailed description of datasets with in-depth dataset statistics in \Cref{tab:data-details}.

\begin{table}
\resizebox{\columnwidth}{!}{%
\setlength{\tabcolsep}{13pt}
\centering
\begin{tabular}{lccc}
\toprule
\multicolumn{1}{c}{\multirow{2}{*}{Data}} & \multicolumn{1}{c}{Task} & \multicolumn{1}{c}{Demog.} & \multicolumn{1}{c}{Demog.} \\ 
     & \multicolumn{1}{c}{classes} & \multicolumn{1}{c}{attributes} & \multicolumn{1}{c}{groups} \\

\specialrule{.4pt}{2pt}{0.2pt}
\rowcolor[HTML]{EFEFEF} \multicolumn{4}{c}{\texttt{ Clinical notes}} \\
In-hosp. Mort. & 2 & gender & 2 \\
Phenotyping & 28 & gender & 2 \\ 
\rowcolor[HTML]{EFEFEF} \multicolumn{4}{c}{\texttt{ Online reviews}} \\
Sentiment & 3 & gender + age & 4 \\
Topic & 8 & gender + age & 4 \\ 
\rowcolor[HTML]{EFEFEF} \multicolumn{4}{c}{\texttt{ Twitter}} \\
Sentiment & 2 & race & 2 \\
HateXplain & 2 & race & 5 \\ 
\bottomrule
\end{tabular}}%
\caption{\label{tab:data-summary}Datasets used in our experiments.}
\end{table}

\subsection{Clinical Records}

We use the Multiparameter Intelligence Monitoring in Intensive Care (MIMIC-III) dataset \citep{johnson2016mimic, johnson2016mimicdatabase, goldberger2000physiobank}, a collection of anonymized English medical records that include clinical notes drawn from a critical care unit at Beth Israel Deaconess Medical Center between 2001 and 2012. 
We select two tasks from those defined by \citet{zhang2020hurtful}:

\textbf{In-hospital Mortality.}
The task is to predict whether a patient will die in the hospital based on the textual content of all the clinical notes created within the first 48 hours of the hospital stay. 

\textbf{Phenotyping.}\footnote{In a medical record, a phenotype is a clinical condition or characteristic.}
The task of assigning medical conditions based on the evidence in the clinical record.
In our task, we will assign up to 25 acute or chronic conditions from the HCUP CCS code groups \citep{harutyunyan2019multitask}, labeled with ICD-9 codes, and three extra summary-labels: any, chronic, or acute condition.
Therefore, the task is modeled as a set of 28 binary classification tasks, and evaluated as a multi-label problem.
We use the same pre-processing pipeline and train-dev-test splits as \citet{zhang2020hurtful}.\footnote{\url{https://github.com/MLforHealth/HurtfulWords}}

\subsection{Online Reviews}

We use the Trustpilot data of \citet{hovy-2015-demographic}: English language reviews of products, stores, and services from an open review platform with a 5-point rating.
For our experiments, we utilize the \texttt{sentiment} (100k reviews) and \texttt{topic} (24k reviews) tasks which share demographics for age -- under 35 (U35) and over 45 (O45) years old -- and gender -- men and women. 

\textbf{Reviews sentiment.} 
Labels assigned based on the stars of the reviews and selected reviews that have both age and gender labels available.

\textbf{Reviews topic.}
Labels assigned based on the general topic of the review, e.g. fashion, fitness, etc. using the Trustpilot taxonomy for seller companies and
selected from the top 5 most popular topics: Fitness \& Nutrition (\textit{Fitness}), Fashion Accessories (\textit{Fashion}), Gaming (\textit{Gaming}), Cell phone accessories (\textit{Cell Phone}) and Hotels (\textit{Hotels})), following \citet{hovy-2015-demographic}.
We perform the same demographic selection criteria as the \textit{sentiment} task.
We obtain randomly stratified train-dev-test (60-20-20\%) splits ensuring equal representations for both gender and age groups.

\subsection{Social Media}

\textbf{Twitter sentiment.}
We use the Twitter sentiment classification task introduced by \citet{elazar-goldberg-2018-adversarial}. Labels were assigned based on common emojis and demographic variables are based on the dialectal corpus from \citet{blodgett-etal-2016-demographic}, where race was assigned based on geolocation and words used in the tweet, obtaining a binary AAE (African-American English) and SAE (Standard American English) which we use as proxies for non-Hispanic African-Americans and non-Hispanic Caucasians.

\textbf{HateXplain.}
A hate speech classification dataset of Twitter and Gab posts \citep{mathew2021hatexplain}.
We use the binary version of the task which classifies for toxicity of posts.
We select the posts for which there is a majority agreement of annotators for race target groups, and for which we have representation across train-dev-test splits.

For each dataset, we follow the splits provided by \citet{elazar-goldberg-2018-adversarial} and \citet{mathew2021hatexplain}, respectively.

\section{Experiments}
\label{sec:experiments}

This section describes baselines and model training.
\cref{tab:experiments} in \cref{apx:experiment-table} shows all combinations of models, training datasets, and fairness attributes.

\subsection{Models}
We implement our fairness objectives in an MTL setting based on a shared language encoder and task-specific classification heads. 
We use BERT-style encoders \citep{devlin-etal-2019-bert} with a domain-specific vocabulary: SciBERT for clinical tasks, pretrained on scientific text \citep{beltagy-etal-2019-scibert}, following prior work \citep{zhang2020hurtful, amir-etal-2021-impact},\footnote{\url{https://huggingface.co/allenai/scibert_scivocab_uncased}} RoBERTa for the online reviews tasks \citep{DBLP:journals/corr/abs-1907-11692}
initialized with the \texttt{roberta-base} checkpoint,\footnote{\url{https://huggingface.co/roberta-base}}
and BERTweet for the social media tasks \citep{nguyen-etal-2020-bertweet}, initialized with the \texttt{vinai/bertweet-base} checkpoint.\footnote{\url{https://huggingface.co/vinai/bertweet-base}}
We add a separate linear classification head for each task, with a Softmax output function to allow for multi-class classification or a Sigmoid output function for binary and multi-label classification. 
The document representation for the classification head is a mean-pooled aggregation across all subword representations of the document taken at the top layer of the network.
The training objective is an additive combination of the loss for each of the individual tasks.
Models were trained on Nvidia A100 GPUs, using \texttt{jiant} \citep{phang2020jiant}, a multi-task wrapper library.

Fairness methods require a careful tradeoff between the task loss and fairness loss \citep{islam2021can}. To obtain the best performing model, we use a grid search for each task, with a learning rate$=[1e^{-4}, 1e^{-5}, 1e^{-6}]$ with Adam optimizer \citep{kingma2014adam}, and batch size$=[16, 32, 48]$. We select the best performing model on development data and report test data results.

\subsection{Baselines}
\label{sec:experiments-baselines}
\label{sec:experiments-fair-baselines}
\label{sec:experiments-mtl-baselines}

We establish baselines against which to compare our MTL fairness transfer method.

\textbf{STL-base.}
We train a single-task model for each task, i.e. a fine-tuned encoder and classification layer.
These models do not include a fairness loss since they represent the classifiers obtained when no demographic attributes are available.
We named these models single task learning base (\texttt{STL-base}), and they serve as an upper bound in task performance when fairness is not a goal.

\textbf{STL-fair.} Finetuning models without fairness losses can result in unfair classifiers \citep{Lan2017DiscriminatoryT, zhang2020hurtful}, which is known as \textit{no fairness through unawareness} \citep{barocas-hardt-narayanan}.
To determine how well we could do in the theoretical with full demographic information, we train single-task models with both a task loss and fairness loss \S\ref{sec:experiments-baselines}.
For the models trained on the clinical dataset and Twitter datasets, we add a single-attribute fairness loss, with gender and race groups respectively.
For the models trained on the online reviews datasets (sentiment and topic), we add an intersectional fairness loss, with age and gender attributes.
This allows us to test both single-attribute and intersectional fairness. We call these single task models with fairness objectives \texttt{STL-fair}.
We performed a grid search on each task, with the same search spaces as before, in addition to the fair-related hyperparameters $\lambda=[.01, .05, .1]$, $\rho=[.01, .1, .9]$, and \textit{burn-in}$=[.5, 1]$ epochs, defined in \S \ref{sec:methods-fair-loss}.

\textbf{MTL-base.} We next evaluate models trained in a multi-task setting. 
While MTL can lead to better performance, it often leads to worse results compared to single-task baselines due to task conflict and other optimization challenges \citep{weller-etal-2022-use, gottumukkala-etal-2020-dynamic}.
A \textit{dynamic scheduler}, which changes the rate that a task is seen based on the current relative performance, has been shown to improve performance in traditional MTL setups \citep{gottumukkala-etal-2020-dynamic}.
Therefore, we first train MTL models with a dynamic scheduler on mutually related task pairs to avoid a domain mismatch: \textit{In-hospital Mortality} \&\textit{Phenotyping} (clinical setting), \textit{reviews sentiment} \& \textit{reviews topic} (online reviews domain), and \textit{Twitter sentiment} \& \textit{HateXplain} (social media setting).
We name these models multi-task baselines \texttt{MTL-base}.

\textbf{BLIND.}
We also compare our work with other bias removal methods that do not require demographic attributes.
\citet{orgad2023blind} propose that often classifiers make predictable mistakes when implicit demographic features are used as shorcut features, a bias also known as \textit{simplicity} bias \citep{bell2023simplicity}.
\textit{BLIND} trains a success classifier that takes the encoder features and predicts the success of the model on the task. 
A correct prediction by the success classifier means the model used a shallow, or simple, decision and the sample is down-weighted.
We use their algorithm implementation\footnote{code: \url{https://github.com/technion-cs-nlp/BLIND}} and perform a hyperparameter search, $\gamma=[1, 2, 4, 8, 16]$, temp$=[1, 2, 4, 8, 16]$, as suggested by authors \citep{orgad2023blind}.
BLIND does not support multi-label tasks so we do not report results for the clinical tasks.

\subsection{Our Methods}
\label{sec:experiments-mtl-fair}
We propose variations on multi-task learning with a fairness loss in support of our proposed setup.

\textbf{MTL-fair.} 
We evaluate the fairness loss applied to one of the two tasks for each in-domain task pair: clinical, online reviews, and social media domains.
We call these models with an MTL objective and a fairness loss \texttt{MTL-fair}.
To report a fair comparison, each of the \texttt{MTL-fair} models is compared with the task for which no fairness loss was added, e.g. for the \textit{In-hospital Mortality} task, we compare the \texttt{STL-base} and \texttt{STL-fair} trained on \textit{In-hospital Mortality} data only, the \texttt{MTL-base} trained on \textit{In-hospital Mortality} and \textit{Phenotyping} (without fairness loss), and the \texttt{MTL-fair} trained on \textit{In-hospital Mortality} and \textit{Phenotyping}, with a fairness loss applied to the \textit{Phenotyping} task only.
We performed a grid search with the same base search space as in \S \ref{sec:experiments-baselines}

\textbf{MTL-inter.} To train intersectionally fair models on two tasks for which we have only a single axis of demographic attributes, we use an MTL objective with two different single-axis fairness losses.
We focus on the online reviews datasets, for which we have sufficient demographic data to support this experiment.\footnote{MIMIC has demographic data but is highly skewed, resulting in intersection groups with only a handful of individuals.}
We call these models that use MTL with intersectionally fair losses \texttt{MTL-inter}.

\subsection{Evaluation}

We utilize established evaluation metrics for all datasets.
The clinical datasets are evaluated at the patient level.
We use the aggregation function from \citet{zhang2020hurtful} since clinical notes are too long to fit in the context window of models; see \S\ref{apx:data-details} for more details.
We report macro-averaged F1 scores for task performance and $\epsilon$-DEO for fairness.
The best model criteria for \texttt{STL-base}, \texttt{MTL-base} and \texttt{BLIND} models is their F1 validation score.
We choose \texttt{STL-fair}, \texttt{MTL-fair} \& \texttt{MTL-inter} models with the lowest $\epsilon$-DEO and at least 95\% performance of the \texttt{STL-base} models on validation.

So far, it has been assumed that there is an extra dataset that has access to demographic attributes within the same domain.
However, due to the scarcity of NLP datasets with access to demographics, it may not be possible to find an eligible dataset within the same domain.
To evaluate the robustness of our method, we test the impact of domain mismatch and task similarity on the MTL models with fairness loss.
We focus on the \textit{Twitter sentiment} task, as it allows us to pair it with a task within the same domain (\textit{HateXplain}), a similar task but in a different domain (\textit{reviews sentiment}) and other tasks with varied domains and task similarities.

\section{Results \& Analysis}

\begin{table}
\centering
\resizebox{\columnwidth}{!}{
\begin{tabular}{rcccc}
\toprule
                    & \multicolumn{4}{c}{Clinical}                                     \\
\cmidrule(lr){2-5} 
                    & \multicolumn{2}{c}{In-hosp Mort.} & \multicolumn{2}{c}{Phenotyping} \\
\cmidrule(lr){2-3} \cmidrule(lr){4-5} 
             & F1 (\%) $\uparrow$    & $\epsilon$-DEO $\downarrow$   & F1 (\%) $\uparrow$    & $\epsilon$-DEO $\downarrow$  \\
 \cmidrule(lr){2-2} \cmidrule(lr){3-3} \cmidrule(lr){4-4} \cmidrule(lr){5-5}  
             
\texttt{STL-base}            & 62.1                  & 0.25                          & \textbf{53.6}         & 0.28                          \\
\texttt{STL-fair}            & 65.1                  & 0.22                          & 52.9                  & 0.26                          \\
\texttt{MTL-base}            & \textbf{65.6}         & \textbf{0.17}                 & 53.3                  & 0.27                          \\
\cmidrule(lr){2-5}
\texttt{MTL-fair}           & 64.0                  & 0.19                          & 53.0                  & \textbf{0.21}                 \\
\midrule
    & \multicolumn{4}{c}{Twitter}                                    \\
    \cmidrule(lr){2-5}
    & \multicolumn{2}{c}{HateXplain} & \multicolumn{2}{c}{Sentiment} \\
    \cmidrule(lr){2-3} \cmidrule(lr){4-5}
     & F1 (\%) $\uparrow$    & $\epsilon$-DEO $\downarrow$   & F1 (\%) $\uparrow$    & $\epsilon$-DEO $\downarrow$   \\
      \cmidrule(lr){2-2} \cmidrule(lr){3-3} \cmidrule(lr){4-4} \cmidrule(lr){5-5}
\texttt{BLIND}              & 70.4                  & 1.15                          & \textbf{77.6}         & 0.30                          \\
\texttt{STL-base}           & 71.3                  & 1.58                          & 76.4                  & 0.33                          \\
\texttt{STL-fair}           & \textbf{71.5}         & 1.63                          & 76.5                  & 0.28                          \\
\texttt{MTL-base}           & 69.9                  & 1.45                          & 76.2                  & 0.37                          \\
\cmidrule(lr){2-5}
\texttt{MTL-fair}           & 70.4                  & \textbf{0.80}                 & 75.5                  & \textbf{0.28}                 \\
\bottomrule
\end{tabular}}
\caption{\label{tab:main-results}
Scores of the MTL fairness loss (\texttt{MTL-fair}) within-domain experiments. Best per task is \textbf{bold}.
}
\end{table}

\cref{tab:main-results} reports performance and fairness scores for within-domain \texttt{MTL-fair} experiments.  
Our baselines perform comparably with prior work \citep{zhang2020hurtful, hovy-2015-demographic, mathew2021hatexplain, elazar-goldberg-2018-adversarial}
so we can evaluate the use of multi-task learning methods to debias algorithms with high-performing models.
In contrast to the common perception that we must trade off fairness and performance, we observe that the performance of \texttt{STL-fair} models is equal to or better in 3/4 tasks compared to the \texttt{STL-base} model baselines and produces fairer models based on $\epsilon$-DEO.
This confirms recent work suggesting that an extensive grid search of hyperparameters avoids the fairness vs. performance trade-off \citep{islam2021can}.\looseness=-1

\textbf{Multi-task fairness generalizes to tasks without demographics.}
We expected the \texttt{STL-fair} models to be an upper bound for fairness, and  \texttt{STL-base} an upper bound for performance compared to the \texttt{MTL-fair} models.
However, for 3/4 tasks, the \texttt{MTL-fair} models are fairer than the \texttt{STL-fair} counterparts!
In these cases, the performance of the \texttt{MTL-fair} models is slightly worse than \texttt{STL-fair} models but still comparable to \texttt{STL-base}, obtaining models that are fairer while maintaining model performance.
This suggests that just as multi-task learning finds representations that are useful for training multiple tasks, multi-task fairness learning corrects model representations to be fairer for both tasks -- sometimes finding a fairness minimum that is fairer than it would with access to target task demographic attributes.
This technique may be yielding more generalizable and fair representations.
Comparing to \texttt{BLIND}, we observe that \texttt{BLIND} yields fairer models than \texttt{STL-base} but less fair than \texttt{STL-fair} and our method \texttt{MTL-fair}.
This suggests that when we have no demographic attributes, \texttt{BLIND} is better than not attempting fairness, but effectively using demographics, whether internally or in another task, increases the fairness of the models.
In all settings, the multi-task fairness loss produced a model that is fairer than the single-task baseline without demographic attributes and with comparable performance.\looseness=-1

\begin{table*}
\centering
\resizebox{2\columnwidth}{!}{
\begin{tabular}{rcccccccccccc}
\toprule
                             & \multicolumn{6}{c}{Reviews sentiment}        & \multicolumn{6}{c}{Reviews topic}           \\
\cmidrule(lr){2-7} \cmidrule(lr){8-13} 
                             &                      &                               &  \multicolumn{4}{c}{F1 (\%) per sub-group $\uparrow$} &                      &                               & \multicolumn{4}{c}{F1 (\%) per sub-group $\uparrow$} \\
\cmidrule(lr){4-7} \cmidrule(lr){10-13} 
                             & F1 (\%) $\uparrow$   & $\epsilon$-DEO $\downarrow$   & \gr F-U35   & \gr F-O45 & \gr M-U35 & \gr M-O45  & F1 (\%) $\uparrow$       & $\epsilon$-DEO $\downarrow$   &\gr F-U35 & \gr F-O45  & \gr M-U35 & \gr M-O45   \\
 \cmidrule(lr){2-2} \cmidrule(lr){3-3} \cmidrule(lr){4-4} \cmidrule(lr){5-5}  \cmidrule(lr){6-6} \cmidrule(lr){7-7} \cmidrule(lr){8-8}  \cmidrule(lr){9-9} \cmidrule(lr){10-10} \cmidrule(lr){11-11} \cmidrule(lr){12-12} \cmidrule(lr){13-13} 
\texttt{BLIND}               & 84.3             & 1.16          & \gr82.7          & \gr\textbf{85.7} & \gr84.4          & \gr83.8          & 92.0          & 1.05          & \gr\textbf{91.7} & \gr86.7          & \gr89.7          & \gr\textbf{89.9} \\
\texttt{STL-base}            & 84.5             & 0.95          & \gr\textbf{87.1} & \gr83.9          & \gr83.1          & \gr84.6          & 91.9          & 1.42          &\gr 90.0          & \gr85.7          & \gr\textbf{90.3} &\gr 88.5          \\
\texttt{STL-fair}            & \textbf{85.6}    & 0.77          & \gr86.4          & \gr84.8          & \gr\textbf{84.6} & \gr\textbf{86.3} & \textbf{92.1} & 1.04          &\gr 90.9          & \gr\textbf{88.7} & \gr90.2          & \gr88.1          \\
\texttt{MTL-base}            & 84.4             & 0.89          & \gr86.1          & \gr84.6          & \gr82.9          & \gr84.7          & 91.6          & 1.52          & \gr91.4          & \gr85.9          & \gr89.4          & \gr89.5          \\
\texttt{MTL-fair}            & 83.6             & 0.65          & \gr85.5          & \gr82.7          & \gr82.8          & \gr83.7          & 91.2          & 0.86          & \gr90.9          & \gr88.3          & \gr88.1          & \gr89.1          \\
\cmidrule(lr){2-7} \cmidrule(lr){8-13}
\texttt{MTL-inter}           & 84.1             & \textbf{0.58} & \gr86.0          & \gr83.7          & \gr82.4          & \gr84.7          & 91.6          & \textbf{0.82} & \gr90.6          & \gr86.6          & \gr89.4          & \gr88.9          \\
\bottomrule
\end{tabular}}
\caption{\label{tab:intersectional-results}
Scores of the intersectional experiments on the reviews datasets (\texttt{MTL-inter}). Best per task is \textbf{bold}.
}
\end{table*}

\textbf{Multi-task enables intersectional fairness.}
\Cref{tab:intersectional-results} shows the results for the intersectional fairness experiments. 
The best \texttt{MTL-inter} model performs comparably to the \texttt{STL-base} and is fairer compared to the \texttt{STL-fair} models in both tasks.
We obtain an intersectionally fairer model compared to the baselines when only one demographic attribute is available per task.
This suggests that the single-attribute fairness losses combine to obtain model representations that are beneficial to the fairness of both protected attributes and their intersectional groups.
Compared to prior work, we see fairness benefits when utilizing single-axis demographics, perhaps due to greater loss stability and the ability of MTL setups to integrate all the losses.

\definecolor{color1}{HTML}{FF9E1B}
\definecolor{color2}{HTML}{0072CE}

\begin{table}
  \resizebox{\columnwidth}{!}{
\begin{tabular}{llcc}
\toprule

 \multicolumn{2}{c}{Method}             & \multicolumn{1}{c}{F1 (\%) $\uparrow$} & \multicolumn{1}{c}{$\epsilon$-DEO $\downarrow$} \\
 \cmidrule(lr){1-2} \cmidrule(lr){3-3}  \cmidrule(lr){4-4} 
\multicolumn{2}{c}{\texttt{BLIND}}                         & \textbf{77.6}        & 0.30                                     \\
\multicolumn{2}{c}{\texttt{STL-base}}                      & 76.4                 & 0.33                                     \\
\multicolumn{2}{c}{\texttt{STL-fair}}                      & 76.5                 & 0.28                             \\
\cmidrule(lr){1-4}  
\texttt{MTL-fair:} & \color{color1}HateXplain   & 75.5                 & 0.28                     \\
& \color{color2}review sentiment    & 76.3                 & \textbf{0.23}                     \\
& \color{gr}review topic                          & 75.7                 & 0.23                     \\
& \color{gr} In-Hosp Mort.                         & 75.8                 & 0.25                     \\
& \color{gr} Phenotyping                           & 75.2                 & 0.32                     \\
                                    
\bottomrule
\end{tabular}}
\caption{Scores of \texttt{MTL-fair} for the Twitter sentiment task paired with different domain and task annotations: \textcolor{color1}{same domain}, \textcolor{color2}{same task}, and \textcolor{gr}{neither}. \textbf{Bold} is best.}
  \label{tab:results-across-domains}
\end{table}

\textbf{Multi-task fairness generalizes across domains and tasks.}
So far we have assumed access to a task with demographic attributes available within the same domain, exploiting text similarities between the tasks to generalize the fairness across tasks.
However, given the scarcity of datasets with demographic attributes, we may wonder whether domain similarity is necessary to transfer fairness. 
In \cref{tab:results-across-domains} we show the results of the single-task \textit{Twitter sentiment} models as well as applying the MTL fair loss across different datasets.
We observe that adding a fairness loss to the MTL settings helps in fairness with tasks across domains and task similarities, except for the clinical \textit{Phenotyping} task.
This may be because the performance of the \textit{Phenotyping} task in the MTL system was poor (possibly because of task incompatibility) and the fairness loss might not have actually provided any meaningful change to the model.
Regardless, on tasks where we obtain competitive performance for both tasks, the fairness loss was able to generalize fairness, obtaining models that are fairer than the single-task baselines and sometimes fairer than applying a fairness loss to the target task, showing evidence that our method is robust across domains, demographic attributes, and task similarities.

\textbf{Why does the multi-task fairness loss work?}
The results in this section suggest that the multi-task fairness loss produces more generalizable and fairer representations.
We hypothesize that the combination of (A) the regularizing effect of the fairness loss, as suggested by prior work \citep{islam2021can}, (B) shared parameters across tasks and (C) the simultaneous learning of both tasks allows for positive fairness transfer.
First, we note that multi-task learning alone (B \& C, \texttt{MTL-base}) or a fairness loss (A, \texttt{STL-fair}) may suffer in performance or fairness (or sometimes both) compared to our method.
Further, one could have shared parameters, B, but not train simultaneously by finetuning on individual tasks consecutively rather than simultaneously, a multi-task method also known as STILT \citep{weller-etal-2022-use, phang2018sentence}.
In \Cref{apx:STILT-experiments} we show that when the fairness loss is applied consecutively, rather than simultaneously, the fairness transfer effect is no longer observed.
Thus, the MTL objective plus the shared parameters are instrumental in enabling the positive transfer of the fairness loss from one task to another.

\section{Related Work}
\label{sec:related-work}

Methods that transfer fairness have used external datasets to ensure fairness and MTL \citep{oneto2020learning} or domain-shift transfer methods \citep{chen2022fairness, schrouff2022diagnosing}; however, they often rely on strong assumptions of distribution shifts, limiting their impact with real-world applications \citep{schrouff2022maintaining} or applicability to NLP methods.
In comparison, while our method does not include explicit domain-shift assumptions, it relies on some domain similarities that are well-studied for general multi-task setups \citep{weller-etal-2022-use}.
Another solution to debias models is to use proxy variables or inferred demographics in settings where we lack demographic data. However, these methods are dependent on the accuracy of the demographic inference model \citep{aguirre-etal-2021-gender, bharti2023estimating} or the availability of proxy variables, e.g. names \citep{romanov-etal-2019-whats}.

MTL has become the standard training setting for Large Language Models (LLM)~\citep{devlin-etal-2019-bert, Radford2019LanguageMA, Brown2020LanguageMA}. 
Unfortunately, studies have found that fine-tuning LLMs often results in unfair models, even when starting from a debiased pre-trained encoder \citep{Lan2017DiscriminatoryT, zhang2020hurtful}. 
Instead, they conclude that fairness requires applying debiasing methods in fine-tuning for the task of interest, requiring demographic information for each task.

In our work we use a separation-based group-wise definition of fairness, \textit{equalized odds} \citep{hardt2016equality}, that was adapted to be differentiable and applied to training procedures inspired by the $\epsilon$-Differential Fairness from \citet{foulds2020intersectional}.
However, many other group-wise definitions of fairness may be adapted for other tasks, e.g. \textit{equalized opportunity} \citep{hardt2016equality}, which ensures equal true positive rates (recall) across demographic subgroups.
There is also \textit{adversarial} fairness loss, where an adversary is added in the training procedure to predict the demographic attributes from the output of the task classifier.
This loss also achieves independence of predictions and demographic attributes, similar to demographic parity, and has found success in similar setups from prior work \citep{islam2021can, zhang2020hurtful}.
Our methods can be easily used with any of these demographic losses in the procedure.

\section{Conclusion}

We explored whether MTL methods for NLP tasks can transfer demographic fairness from one task to another.
To achieve this, we adapted single-task fairness losses to multi-task settings to transfer fairness across tasks.
We tested our method in multiple NLP datasets in different domains: clinical notes \citep{johnson2016mimic, johnson2016mimicdatabase, goldberger2000physiobank}, online reviews \citep{hovy-2015-demographic} and social media \citep{mathew2021hatexplain, elazar-goldberg-2018-adversarial}.
We found that while MTL alone and other consecutive variations of MTL (e.g. STILTS) do not help in fairness and may hurt performance, MTL methods with our fairness loss are able to debias models using the demographic attributes from a secondary task, opening up the possibility for producing fair models for a wide range of tasks that lack demographic data.
This finding also informs future work on MTL, suggesting adding regularizers, e.g. fairness losses, can help in performance deficits found in prior work \citep{weller-etal-2022-use, gottumukkala-etal-2020-dynamic}.

Additionally, we showed that MTL methods can debias models for intersectional fairness by leveraging two tasks, each with different demographic attributes, to learn a model that achieves intersectional fairness on both tasks.
This finding opens up the integration of intersectional fairness losses to new applications and settings that were previously restricted by limited access to demographic attributes.
Finally, we test the ability of the MTL fairness loss to generalize fairness across domains and tasks, we find that the transfer of fairness is not dependent on domain or task similarity, but rather related to the performance of the secondary task.
Our methods increase the range of tasks that fairness methods can be applied to in the machine learning and NLP community, by allowing the use of external tasks that have demographic attributes to obtain fairer models.

\section{Limitations}

Our results suggest that our MTL methods are able to utilize external demographic attributes to achieve better fairness for our target task.
However, the selection criteria for the best-performing models require access to demographic attributes for the test set to assess the fairness of the models.
A solution to this would be to select the models that are the best performing for our target task with the lowest fairness score for the task that we do have demographic data available.
This selection criteria, however, does not guarantee the most optimal model, especially if the demographic attribute distributions or the task domains are different.
Our recommendation is to validate the fairness of the models with access to demographic attributes when possible.

\section{Ethics Statement}

We address intersectionality as intersectional group fairness in the methods and analysis when possible given the data availability, as they enable a practical approach for inquiry of these models.
We acknowledge that there are real interlocking systems of power that contribute to causing these disparities in society, and that our dataset capture these.
For example, we evaluate models on the clinical domain using the MIMIC-III dataset: the healthcare system has been historically biased against people in groups in many protected attribute axis e.g. socio-economic status, race/ethnicity, gender,  and age.
The goal of our approach is to address these biases in machine learning models so they are less likely to exacerbate the real-life biases as they are integrated in society.


\bibliography{anthology,custom}
\bibliographystyle{acl_natbib}
\clearpage
\newpage
\appendix
\label{sec:appendix}
\renewcommand{\thepage}{}

\section{Fairness Definition}
\label{apx:fairness-definition}

$\epsilon$-Differential Fairness is a demographic-parity based metric, which requires that the demographic attributes are \textit{independent} of the classifier output \citep{barocas-hardt-narayanan, foulds2020intersectional}.
Formally, we assume a finite dataset of size $N$, with each sample consisting of three attributes: features $x$ (in our datasets these are text sequences), task labels $y$, and demographic attributes $z$.
Let $s_1, ... , s_p$ be discrete-valued demographic attributes, $z = s_1 \times s_2 \times ... \times s_p$. A model $M(X)$ satisfies $\epsilon$-DF with respect to $z$ if for all $x$, and $\hat{y} \in Range(M)$,
\begin{equation*}
    e^{-\epsilon} \leq \frac{Pr(M(x) = \hat{y} | \zeta_i)}{Pr(M(x) = \hat{y} | \zeta_j)} \leq e^{\epsilon},
\end{equation*}
for all $(\zeta_i, \zeta_j) \in z \times z$ where $Pr(\zeta_i) > 0$, $Pr(\zeta_j) > 0$. Smaller $\epsilon$ is better with $\epsilon$ = 0 meaning perfect fairness \citep{foulds2020intersectional}.
Perfect fairness under this definition means that the rates of predicted labels are the same across demographic groups, achieving independence between demographic attributes and predictions.

In short, $\epsilon$-Differential Fairness is an independence-based metric that measures the biggest difference in prediction rates between intersections of demographic attributes.
However, independence based fairness definitions, like demographic parity and $\epsilon$-DF, have limitations in settings where the prevalence of the target labels is somehow related to the demographic attributes, e.g. breast cancer is much more common in women than men.
In these settings, independence based definitions would require model predictions to be independent of the demographic attributes, which would encourage lower performance on the desired task, e.g. either an increase in the prediction of breast cancer for men and/or a decrease in breast cancer for women which are both not ideal.
For these reasons, we favor a separation based metric, like \textit{equalized odds}, that avoids limitations associated with dependence of model predictions on demographics by requiring independence conditioned on the target variable \citep{hardt2016equality}, i.e. that both recall and specificity rates are equal across demographic groups.

We apply \textit{equalized odds} on the $\epsilon$-DF framework to obtain a metric that is also differentiable, and call it  $\epsilon$-Differential Equalized Odds ($\epsilon$-DEO).
Formally, let $s_1, ... , s_p$ be discrete-valued demographic attributes, and  $z = s_1 \times s_2 \times ... \times s_p$ the intersectional groups. A model $M(X)$ satisfies $\epsilon$-DEO with respect to $z$ if for all $x$, $\hat{y} \in \textrm{Range}(M)$ and $y \in \textrm{Range}(M)$,
\begin{equation}
    e^{-\epsilon} \leq \frac{Pr(M(x) = \hat{y} | \zeta_i, y)}{Pr(M(x) = \hat{y} | \zeta_j, y)} \leq e^{\epsilon},
\end{equation}
for all $(\zeta_i, \zeta_j) \in z \times z$ where $Pr(\zeta_i) > 0$, $Pr(\zeta_j) > 0$; smaller $\epsilon$ is better, with $\epsilon$ = 0 for perfect fairness.
Perfect fairness results from a classifier with the same recall and specificity rates across intersectional groups of demographic attributes.

\section{STILT and frozen experiments}
\label{apx:STILT-experiments}
In this section we test the hypothesis of whether it is important to have shared parameters and simultaneous learning when implementing the multi-task fairness loss.

\textbf{MTL.}
We label \texttt{MTL} the models that were trained simultaneously, as described in \S \ref{sec:methods-mtl-fairness}.

\textbf{STILT.}
We label \texttt{STILT} the models that were trained consecutively.
First, the model is finetuned only for task B  with the fairness loss, the task with demographic attributes as seen in \Cref{fig:MTL-fair}.
This step results in a model similar to \texttt{STL-fair} for task B.
Second, the model is further finetuned for task A (as seen in \Cref{fig:MTL-fair}), with a different classification layer and without a fairness loss.
Both steps together result in a model that has been trained with the same data and the same number of parameters as \texttt{MTL-fair}, however the tasks are not trained simultaneously.

\textbf{Frozen.}
In order to test the importance of parameter sharing, we train a variance of the model where the shared parameters, BERT-based encoder, are frozen during training.
In this way, the number of shared parameters, $\theta_s$ in \Cref{tab:fair-losses}, is empty.
First, we train a single-task model with a fairness loss where the encoder is frozen, we label this \texttt{STL-fair-frozen}.
We also train a STILT model, where we first finetune for the task that has demographic attributes (Task B) with a fairness loss end-to-end, and then we finetune for the task without demographic attributes without a fairness loss and with the encoder frozen. 
The idea is that the fairness loss will influence the encoder towards a fairer minima that then the classification loss for the second task will be able to exploit.

\begin{table}
 \centering
 
\begin{tabular}{l@{\hskip -10mm}lccc}
\toprule
          &        &  \multicolumn{1}{c}{F1 (\%) $\uparrow$} & \multicolumn{1}{c}{$\epsilon$-DEO $\downarrow$}\\
          \cmidrule(lr){3-3} \cmidrule(lr){4-4} \cmidrule(lr){5-5}
\texttt{STL-base}  &    & 71.3                   & 1.58                     \\
\texttt{BLIND}     &    & 70.4                   & 1.15                     \\
\texttt{STL-fair}  &    & \textbf{71.5}          & 1.63                     \\
    &\texttt{-frozen}   & 61.8                   & 0.69                     \\
\texttt{STILT-fair} &   & 70.4                   & 1.42                     \\
 &\texttt{-frozen}      & 63.4                   & \textbf{0.60}            \\
 \cmidrule(lr){3-4}
 \texttt{MTL-fair}   &   & 70.4                   & 0.80                     \\
 \bottomrule
\end{tabular}
\caption{Scores for the \texttt{STILT} and \texttt{frozen} version of the model on HateXplain dataset.}
  \label{tab:STILT-results}
\end{table}

\Cref{tab:STILT-results} shows the results for \texttt{STILT-fair}, and the frozen versions \texttt{STL-fair-frozen} and \texttt{STILT-fair-frozen}.
First we see that the frozen versions of the models drastically underperform compared to the end-to-end models ($\Delta$F1 $\approx 10$.) while also being more fair.
This is a clear example of the accuracy-fairness trade-off, which is expected given the drastically smaller amount of parameters available for training for these frozen models.
It is clear that these models are fairer because they perform equally worse for all demographic groups.

When comparing the \texttt{STILT-fair} to our method \texttt{MTL-fair}, we see that while the performance of the models is very similar (both scoring $70.4$ F1), the fairness is drastically better in the simultaneous training (\texttt{MTL-fair} $\epsilon$-DEO=$.80$) vs. consecutively (\texttt{STILT-fair} $\epsilon$-DEO=$1.42$).
This suggests that the MTL objective, which allows for both tasks to influence the learning, is instrumental for the fairness loss on task B to transfer to task A.

\section{Data Details}
\label{apx:data-details}

\begin{table}[]
    \centering
    \resizebox{\columnwidth}{!}{
    \begin{tabular}{l@{\hskip -15mm}lccc}
    \toprule
                  &                  & train  & val   & test  \\
\cmidrule(lr){3-3} \cmidrule(lr){4-4} \cmidrule(lr){5-5}
In-Hosp Mort.     &                  & 13191  & 2701  & 2445  \\
\\
                  & Men              & 55.4   & 54.8  & 55.2  \\
                  & Women            & 44.6   & 45.2  & 44.8  \\
                  \cmidrule(lr){3-5}
                  & Positive         & 13.1   & 13.8  & 11.5  \\
                  \cmidrule(lr){1-5}
Phenotyping       &                  & 13839  & 2850  & 2519  \\
\\
                  & Men              & 57.2   & 55.8  & 56.4  \\
                  & Women            & 42.8   & 44.2  & 43.6  \\
                  \cmidrule(lr){3-5}
                  & Upper Resp.      & 2.6    & 2.5   & 2.6   \\
                  & Lower Resp.      & 3.5    & 4.0   & 3.7   \\
                  & Shock            & 3.8    & 3.6   & 4.2   \\
                  & Any Acute        & 70.8   & 69.9  & 70.6  \\
                  & Any Chronic      & 77.1   & 78.5  & 76.8  \\
                  & Any Disease      & 89.6   & 90.6  & 90.1  \\
                  \cmidrule(lr){1-5}
reviews sentiment &                  & 58259  & 19420 & 19420 \\
\\
                  & Men Under 35     & 23.2   & 23.2  & 23.2  \\
                  & Men Over 45      & 34.7   & 34.7  & 34.7  \\
                  & Women Under 35   & 14.8   & 14.8  & 14.7  \\
                  & Women Over 45    & 27.3   & 27.3  & 27.3  \\
                  \cmidrule(lr){3-5}
                  & positive         & 84.5   & 84.5  & 84.5  \\
                  & neutral          & 3.5    & 3.5   & 3.5   \\
                  & negative         & 12.0   & 12.0  & 12.0  \\
                  \cmidrule(lr){1-5}
reviews topic     &                  & 14744  & 4915  & 4915  \\
\\
                  & Men Under 35     & 54.0   & 54.0  & 54.0  \\
                  & Men Over 45      & 14.2   & 14.2  & 14.3  \\
                  & Women Under 35   & 21.1   & 21.1  & 21.1  \\
                  & Women Over 45    & 10.7   & 10.7  & 10.6  \\
                  \cmidrule(lr){3-5}
                  & Fitness          & 39.6   & 39.5  & 39.6  \\
                  & Fashion          & 16.6   & 16.6  & 16.7  \\
                  & Gaming           & 16.0   & 16.0  & 16.0  \\
                  & Cell Phone       & 14.4   & 14.4  & 14.4  \\
                  & Hotels           & 13.4   & 13.4  & 13.4  \\
                  \cmidrule(lr){1-5}
HateXplain        &                  & 5376   & 661   & 681   \\
\\
                  & African          & 54.5   & 54.0  & 55.1  \\
                  & Arab             & 18.8   & 18.8  & 17.8  \\
                  & Asian            & 6.2    & 6.2   & 6.5   \\
                  & Hispanic         & 5.4    & 5.1   & 5.1   \\
                  & Caucasian        & 15.1   & 15.9  & 15.6  \\
                  \cmidrule(lr){3-5}
                  & Toxic            & 81.3   & 81.2  & 79.7  \\
                  \cmidrule(lr){1-5}
twitter sentiment &                  & 156000 & 4000  & 8000  \\
\\
                  & African American & 50.0   & 50.0  & 50.0  \\
                  & Caucasian        & 50.0   & 50.0  & 50.0  \\
                  \cmidrule(lr){3-5}
                  & Happy            & 50.0   & 50.0  & 50.0  \\
                  & Sad              & 50.0   & 50.0  & 50.0  \\
                  \bottomrule
\end{tabular}}
    \caption{Total (first line) and percentage of documents in the  splits all the datasets, separated by demographics and then task labels.}
    \label{tab:data-details}
\end{table}

In this section, we report dataset statistics, including the number of posts per label and demographic.
We select datasets in varied domains: clinical text records, online reviews, and social media, with both single and intersectional demographic attributes, gender, race and gender+age subgroups, and in a variety of classification paradigms: multiclass, binary and multilabel.
\Cref{tab:data-details} shows the total and percentage for all datasets.

\subsection{Clinical Records}
It is crucial to implement behavioral fairness measures to secure fair behavior in the critical context of AI applications for medical records.
We use the Multiparameter Intelligence Monitoring in Intensive Care (MIMIC-III) dataset \citep{johnson2016mimic, johnson2016mimicdatabase, goldberger2000physiobank}, a collection of anonymized English medical records that include clinical notes drawn from a critical care unit from the Beth Israel Deaconess Medical Center between 2001 and 2012. 
We select two tasks from those defined by \citet{zhang2020hurtful}: in-hospital mortality and phenotyping. 
We use the same pre-processing pipeline as \citet{zhang2020hurtful}\footnote{\url{https://github.com/MLforHealth/HurtfulWords}} 
and only use gender demographics since the other attributes are highly imbalanced, resulting in very small subgroups, as noted by prior work \citep{amir-etal-2021-impact}.
These tasks should be evaluated at the patient level \citep{zhang2020hurtful}, however, because the clinical notes are too long to fit in the input size of the encoder, we created subsequences using sliding windows.
The model predicts a label for each subsequence and at evaluation time we aggregate these predictions to obtain a single prediction for each patient. 
We use an aggregation function from prior work \citep{zhang2020hurtful}: 
\begin{equation*}
    Pr(y=1 | \hat{Y}) = \frac{\max(\hat{Y}) + mean(\hat{Y})n/c}{1 + n/c},
\end{equation*}
where $\hat{Y}$ are the predictions for all the subsequences from a patient, $n$ is the number of subsequences and $c$ is a scaling factor ($c=2$ \citep{zhang2020hurtful}.)

\textbf{In-hospital Mortality.}
The task of in-hospital mortality is to predict whether a patient will die in the hospital based on the textual content of all the clinical notes created within the first 48 hours of the hospital stay. 
To avoid low information notes, we limit the notes to ``nurse", ``nursing/other" and ``physician" types.
We concatenate all notes available within the specified time period and tokenize the concatenated notes and split them into sliding subsequences of 512 subwords, to fit within the BERT context window~\citep{devlin-etal-2019-bert}. We limit the number of subsequences per patient by selecting the last 30 subsequences of the concatenated notes, following \citet{zhang2020hurtful}.

\textbf{Phenotyping.}
In a medical record, a phenotype is a clinical condition or characteristic. Phenotyping is the task of assigning these conditions based on the evidence in the medical record.
In our task, we will assign up to 25 acute or chronic conditions from the HCUP CCS code groups \citep{harutyunyan2019multitask}, labeled with ICD-9 codes.
In addition to those conditions, three summary labels are also added for patients that have any chronic or acute condition.
Therefore, the task is modeled as a set of 28 binary classification tasks, and evaluated as a multi-label problem.
For this task we select the first note written by a ``nurse", ``nursing/other" or ``physician" within the first 48 hours of the stay, as proposed by \citet{zhang2020hurtful}.

For each dataset, we use the train-dev-test splits provided by \citet{zhang2020hurtful}.
\cref{tab:data-details} shows the final breakdown of the number of subsequences in the datasets.

\subsection{Online Reviews}
Developing automated NLP methods for online product reviews can help companies understand customer feedback, improve the user experience, and enable market analysis. There are a variety of tasks defined for online reviews, such as sentiment analysis, determining the helpfulness of a review, and the topic of the review. Furthermore, reviews are authored by a diverse population and we seek models that perform fairly across this user population.

We use data from Trustpilot, an open review platform that allows users to review a range of products, stores, and services \citep{hovy-2015-demographic}.
Each instance is an English language review selected from the Trustpilot website that consists of a text review and a 5-point star rating, along with item information, such as the seller.
The original dataset defined three tasks: \texttt{sentiment} (based on the rating of the review), \texttt{topic} (the subject of the review), and \texttt{attributes} (demographic attributes of the review author).
For our experiments, we utilize the \texttt{sentiment} (100k reviews) and \texttt{topic} (24k reviews) tasks which share demographics for age -- under 35 (U35) and over 45 (O45) years old -- and gender -- men and women.

\textbf{Reviews sentiment.} 
This is a multiclass task where the labels were assigned based on the stars of the reviews: 1-star reviews were labeled as ``negative'', 3-star labeled as ``neutral'' and 5-star labeled as ``positive''.
We selected reviews that have both age and gender labels available with age ranges between 16-35 and 45-70 years old, and discarded reviews with 2 and 4 stars.

\textbf{Reviews topic.}
This is a multiclass task where labels are assigned based on the general topic of the review, e.g. fashion, fitness, etc.
These concepts were assigned to each review using the Trustpilot taxonomy for seller companies, which summarizes the services and products offered by each company in the corpus with high-level concepts.
We selected the top 5 most popular topics: Fitness \& Nutrition (\textit{Fitness}), Fashion Accessories (\textit{Fashion}), Gaming (\textit{Gaming}), Cell phone accessories (\textit{Cell Phone}) and Hotels (\textit{Hotels})).
We perform the same demographic selection criteria as the \texttt{sentiment} task, resulting in a multiclass task with 5 labels.

For each dataset, we obtain randomly stratified train-dev-test (60-20-20\%) splits ensuring equal representations for both gender and age groups.
For each review, we follow prior work \citep{hung-etal-2023-demographic}
and set the maximum sequence length to 512 subword tokens, the max input size of BERT-style models \citep{devlin-etal-2019-bert}.
\cref{tab:data-details} shows the final breakdown of the number of reviews in the datasets.

\subsection{Social Media}
Social media platforms host a diverse population, with studies demonstrating NLP system bias on related tasks \citep{aguirre-etal-2021-gender}.

\textbf{Twitter sentiment.}
This is a binary sentiment classification task using Twitter data.
Sentiment labels were assigned based on common emojis, following the preprocessing procedure of \citet{elazar-goldberg-2018-adversarial}.
The demographic variables are based on the dialectal corpus from  \citet{blodgett-etal-2016-demographic}, where race was assigned based on geolocation and words used in the tweet, obtaining a binary AAE (African-American English) and SAE (Standard American English) which we use as proxies for non-Hispanic African-Americans and non-Hispanic Caucasians.

\textbf{HateXplain.}
This hate speech classification dataset combines Twitter and Gab messages \citep{mathew2021hatexplain}.
We use the binary version of the task which identifies toxicity of posts.
We select the posts for which there is a majority agreement of annotators for race target groups, and for which we have representation across train-dev-test splits.

For each dataset, we follow the splits provided by \citet{elazar-goldberg-2018-adversarial} and \citet{mathew2021hatexplain} respectively.
\cref{tab:data-details} shows the number of posts for the \textit{HateXplain} and \textit{Twitter sentiment} datasets respectively.

\section{Experiment Table}
\label{apx:experiment-table}

For each dataset, the model setup and their respective training data, fairness loss attribute and which task the fairness loss was applied to. \texttt{MTL-fair} are the models with the fairness loss from \S \ref{sec:methods-mtl-fairness}, and \texttt{MTL-inter} is the model with the intersectional fairness loss discussed in \S \ref{sec:methods-mtl-intersectional-fairness}. * The \texttt{MTL-inter} model uses two separate single-attribute fairness losses for each task.

\begin{table*}[t]
  \caption{list of experiments}
  \label{tab:experiments}
  \resizebox{2\columnwidth}{!}{
  \begin{tabular}{llll}
  \toprule
\multicolumn{4}{c}{Review Sentiment}                                                       \\
& training data   & fairness loss attributes & fairness loss target task \\
\cmidrule(lr){2-2} \cmidrule(lr){3-3} \cmidrule(lr){4-4}
STL-base           & sentiment       & no                       & no                        \\
STL-fair           & sentiment       & gender+age               & sentiment                 \\
MTL-base           & sentiment+topic & no                       & no                        \\
MTL-fair           & sentiment+topic & gender+age               & topic                     \\
\midrule
\multicolumn{4}{c}{Review Topic}                                                            \\
& training data   & fairness loss attributes & fairness loss target task \\
\cmidrule(lr){2-2} \cmidrule(lr){3-3} \cmidrule(lr){4-4}
STL-base           & topic           & no                       & no                        \\
STL-fair           & topic           & gender+age               & topic                     \\
MTL-base           & sentiment+topic & no                       & no                        \\
MTL-fair           & sentiment+topic & gender+age               & sentiment                 \\
\midrule
\multicolumn{4}{c}{In-Hospital Mortality}                                                              \\
& training data   & fairness loss attributes & fairness loss target task \\
\cmidrule(lr){2-2} \cmidrule(lr){3-3} \cmidrule(lr){4-4}
STL-base           & In-hosp Mort.            & no                       & no                        \\
STL-fair           & In-hosp Mort.           & gender                   & In-hosp Mort.                     \\
MTL-base           & In-hosp Mort.+Phenotyping     & no                       & no                        \\
MTL-fair           & In-hosp Mort.+Phenotyping     & gender                   & Phenotyping                    \\
\midrule
\multicolumn{4}{c}{Phenotyping}                                                             \\
& training data   & fairness loss attributes & fairness loss target task \\
\cmidrule(lr){2-2} \cmidrule(lr){3-3} \cmidrule(lr){4-4}
STL-base           & Phenotyping          & no                       & no                        \\
STL-fair           & Phenotyping          & gender                   & Phenotyping                    \\
MTL-base           & In-hosp Mort.+Phenotyping     & no                       & no                        \\
MTL-fair           & In-hosp Mort.+Phenotyping     & gender                   & In-hosp Mort.                     \\
\midrule
\multicolumn{4}{c}{Twitter Sentiment}                                                             \\
& training data   & fairness loss attributes & fairness loss target task \\
\cmidrule(lr){2-2} \cmidrule(lr){3-3} \cmidrule(lr){4-4}
STL-base           & Twitter sentiment          & no                       & no                        \\
STL-fair           & Twitter sentiment          & race                   & twitter sentiment                    \\
MTL-base           & HateXplain+Twitter sentiment     & no                       & no                        \\
MTL-fair           & HateXplain+Twitter sentiment     & race                   & HateXplain                     \\
\midrule
\multicolumn{4}{c}{HateXplain}                                                             \\
& training data   & fairness loss attributes & fairness loss target task \\
\cmidrule(lr){2-2} \cmidrule(lr){3-3} \cmidrule(lr){4-4}
STL-base           & HateXplain          & no                       & no                        \\
STL-fair           & HateXplain          & race                   & HateXplain                    \\
MTL-base           & Twitter sentiment+HateXplain     & no                       & no                        \\
MTL-fair           & Twitter sentiment+HateXplain     & race                   & Twitter sentiment                    \\
\midrule
\multicolumn{4}{c}{Intersectional Experiments}                                                          \\
& training data   & fairness loss attributes & fairness loss target task \\
\cmidrule(lr){2-2} \cmidrule(lr){3-3} \cmidrule(lr){4-4}
STL-base-sentiment & sentiment       & no                       & no                        \\
STL-base-topic     & topic           & no                       & no                        \\
STL-fair-sentiment & sentiment       & gender+age               & sentiment                 \\
STL-fair-topic     & topic           & gender+age               & topic                     \\
MTL-base           & sentiment+topic & no                       & no                        \\
MTL-inter       & sentiment+topic & gender/age*              & sentiment/topic* \\
\bottomrule
\end{tabular}}
\end{table*}

\section{Results without access to val set demographic attributes}
The selection criteria for the best-performing models requires access to demographic attributes for the test set of the target task to assess the fairness of the models.
In the absence of this, \cref{tab:mtl-fairness-loss-no-demo} shows the results for the model setting where we select models with the target task performance of at least 95\% of \texttt{STL-base} and with the lowest fairness score of the auxiliary task.
These models are labeled as \texttt{MTL-fair no demo}.
For all of the datasets, \texttt{MTL-fair no demo} are less fair than if we could select models based on the fairness of the target task, \texttt{MTL-fair}.
In some cases, we obtain models that are less fair than our single-task baselines (\texttt{STL-base}, 2/4) and multi-task baselines (\texttt{MTL-base}, 3/4).
This suggest that while we are able to generalize the fairness loss to other tasks during training, the fairness measures across tasks are not related.
For these reasons we recommend that \texttt{MTL-fair} models are validated for fairness on the target task.

\begin{table*}[t]
  \caption{Scores of the multi-task fairness loss experiments. For the Phenotyping task, these are macro-averages over all labels. Bold is best per task.}
  \label{tab:mtl-fairness-loss-no-demo}
  \resizebox{2\columnwidth}{!}{
\begin{tabular}{rrrcccc}
\toprule

 &         & method               & \multicolumn{1}{c}{AUROC (\%) $\uparrow$} & \multicolumn{1}{c}{$\epsilon$-DEO $\downarrow$} & \multicolumn{1}{c}{$\Delta$Recall (\%) $\downarrow$} & \multicolumn{1}{c}{$\Delta$Specificity (\%) $\downarrow$} \\
 \cmidrule(lr){3-3} \cmidrule(lr){4-4} \cmidrule(lr){5-5}  \cmidrule(lr){6-6} \cmidrule(lr){7-7} 
\multirow{8}{*}{clinical} & In-hosp Mort.    & stl-base             & 77.7                    & 0.22                             & 2.05                      & 5.99                          \\
                                            & & stl-fair             & 77.5                    & 0.18                              & 3.46                     & \textbf{3.54}                 \\
                                            & & mtl-base             & \textbf{78.1 }          & 0.17                              & \textbf{0.23}                     & 4.45                 \\
                                            & & mtl-fair             & 78.1                    & \textbf{0.14}                     & 0.98            & 3.83            \\
                                            & & mtl-fair no demo.    & 78.4                    & 0.18                               & 1.80             & 4.02 \\
\cmidrule(lr){3-7}                                           
                            & Phenotyping     & stl-base             & 69.5                     & 0.24                             & 4.97                     & 3.17                        \\
                                            & & stl-fair             & 69.6                    & \textbf{0.21}                    & \textbf{4.63}            & 2.96                 \\
                                            & & mtl-base             & 69.7                    & 0.29                             & 5.47                     & 4.12                          \\
                                            & & mtl-fair             & \textbf{69.9}           & 0.23                             & 5.94                    & \textbf{2.46}      \\  
                                            & & mtl-fair no demo.    & 70.9                     & 0.28                            & 6.18                    & 4.25 \\
\midrule

 &      & method             & \multicolumn{1}{c}{F1 (\%) $\uparrow$} & \multicolumn{1}{c}{$\epsilon$-DEO $\downarrow$} & \multicolumn{1}{c}{$\Delta$F1 (\%) $\downarrow$} &\\
 \cmidrule(lr){3-3} \cmidrule(lr){4-4} \cmidrule(lr){5-5} \cmidrule(lr){6-6} 
 
\multirow{8}{*}{reviews}  & sentiment & stl-base           & 83.9                 & 0.83                             & 3.79                     & \\
                                    & & stl-fair           & \textbf{86.1}       & 0.68                             & 3.05                        &\\
                                    & & mtl-base           & 83.5                 & 0.66                             & 4.75                        &\\
                                    & & mtl-fair           & 84.4                 & \textbf{0.63}                    & \textbf{1.96}                        &\\
                                    & & mtl-fair no demo.  & 83.3                 & 0.89                              & 5.92 &\\
                                    
\cmidrule(lr){3-6}
                        & topic       & stl-base           & 91.9                 & 1.42                             & 4.58                        &\\
                                    & & stl-fair           & \textbf{92.1}        & 1.04                             & \textbf{2.86}               &\\
                                    & & mtl-base           & 91.3                 & 1.10                             & 6.15                        &\\
                                    & & mtl-fair           & 91.6                 & \textbf{0.85}                    & 3.22                        &\\   
                                    & & mtl-fair no demo.  & 91.3                 & 1.11                             & 4.79                        &\\
\bottomrule
\end{tabular}}
\end{table*}

\end{document}